\documentclass[journal]{IEEEtran}
\usepackage{amsmath,amsfonts}
\usepackage{algorithm}
\usepackage{array}
\usepackage[caption=false,font=normalsize,labelfont=sf,textfont=sf]{subfig}
\usepackage{textcomp}
\usepackage{stfloats}
\usepackage{url}
\usepackage{verbatim}
\usepackage{graphicx}
\usepackage{cite}
\usepackage{graphicx}
\graphicspath{{Figures/}}
\usepackage{algorithm}
\usepackage{algpseudocode}
\usepackage{amsmath}
\usepackage{multirow}
\usepackage{tabularx}
\usepackage{booktabs}
\usepackage{slashbox,multirow}
\usepackage{threeparttable}
\usepackage[colorlinks,linkcolor=blue]{hyperref}

\usepackage{amsmath,amsthm,amssymb,amsfonts}
\usepackage{mathrsfs}

\usepackage{color}
\usepackage[switch]{lineno}

\begin{document}

\title{LBL: Logarithmic Barrier Loss Function for One-class Classification}

\author{Ziyang~Jiang,
        Peng~Lin$^{\dag}$,
        Tianlei~Wang

\thanks{This work was supported by the Zhejiang Provincial Natural Science Foundation of China under Grant No. ZCLMS26F0305. ($^{\dag}$Corresponding author: Peng~Lin)}
\thanks{Z. Jiang, P. Lin, and T. Wang are with Machine Learning and I-health International Cooperation Base of Zhejiang Province, and Artificial Intelligent Institute, Hangzhou Dianzi University, Zhejiang, 310018, China.}
\thanks{E-mail: jiangziyang@hdu.edu.cn, tianleiwang@hdu.edu.cn, penglin@hdu.edu.cn}
}

\markboth{Journal of \LaTeX\ Class Files,~Vol.~14, No.~8, August~2021}%
{Shell \MakeLowercase{\textit{et al.}}: A Sample Article Using IEEEtran.cls for IEEE Journals}


\maketitle

\begin{abstract}
    One-class classification (OCC) aims to train a classifier only with the target class data and attracts great attention for its strong applicability in real-world application. Despite a lot of advances have been made in OCC, it still lacks the effective OCC loss functions for deep learning. In this paper, a novel logarithmic barrier function based OCC loss (LBL) that assigns large gradients to the margin samples and thus derives more compact hypersphere, is first proposed by approximating the OCC objective smoothly. But the optimization of LBL may be instability especially when samples lie on the boundary leading to the infinity loss. To address this issue, then, a unilateral relaxation Sigmoid function is introduced into LBL and a novel OCC loss named LBLSig is proposed. The LBLSig can be seen as the fusion of the mean square error (MSE) and the cross entropy (CE) and the optimization of LBLSig is smoother owing to the unilateral relaxation Sigmoid function. The effectiveness of the proposed LBL and LBLSig is experimentally demonstrated in comparisons with several popular OCC algorithms on different network structures. The source code can be found at \url{https://github.com/ML-HDU/LBL_LBLSig}.
\end{abstract}

\begin{IEEEkeywords}
One-class classification, anomaly detection, logarithmic barrier loss, hypersphere optimization.
\end{IEEEkeywords}

\section{Introduction}\label{Introduction}

One-class classification (OCC), also known as the anomaly detection (AD), aims to construct the model only exploiting the characteristics of the target data. It has attracted much attention in various communities, such as data mining, machine learning and computer vision \cite{review_DL_AD, 9774889}, due to its significance in many applications \cite{9585408, 11208785}. The classical methods, such as the one-class support vector machine (OCSVM) \cite{OCSVM} and the support vector data description (SVDD) \cite{SVDD}, are confronted with difficulties when dealing with high-dimensional and complex data, and such shallow methods are always required feature engineering. In contrast, deep learning-based OCC algorithms achieve exceptional successes, including the autoencoder (AE) based methods \cite{Deep_AD_AE_pidhorskyi2018generative, Deep_AD_AE_zhou2017anomaly, Deep_AD_AE_gong2019memorizing, Deep_AD_AE_wang2020advae, Deep_AD_AE_DAE_reconstruction, Deep_AD_VAE_reconstruction, Deep_AD_ensemble_AEs_reconstruction, OCC_TNNLS_MOCCA_shit, Deep_AD_AE_LSAR, Tianlei_WSI_AE, Deep_AD_SAE_OCSVM, Deep_AD_DAE_OCSVM, Deep_AD_DBN_OCSVM}, the generative adversarial network (GAN) based methods \cite{Deep_AD_GAN_AnoGAN, Deep_AD_GAN_AnoGAN_ALOCC,Deep_AD_GAN_AnoGAN_OGN, Deep_AD_GAN_abati2019latent, Deep_AD_GAN_sabokrou2018adversarially, Deep_AD_GAN_OCGAN, Deep_AD_GAN_schlegl2019f}, and other discrimination-based methods \cite{OCC_NIPS_Deep_Anomaly_Detection_GT, OCC_GODS_OCSVM, Deep_SVDD, OCC_NeurIPS2020_HRN, OCC_IS_OCITN_shit, OCC_CVPR_PANDA_supershit, transformaly_cvpr2022}.
However, little attention is paid to the OCC loss function design. Recently typical OCC loss functions are developed in \cite{Deep_SVDD} and \cite{OCC_NeurIPS2020_HRN}. Ruff \emph{et al.} \cite{Deep_SVDD} propose the soft-boundary loss (SBL) that can be regarded as an extension of the mean square error (MSE) in OCC. But SBL is only demonstrated on the LeNet-type convolution neural network (CNN) and the performance is not promising. In addition, SBL is essentially the hard sample mining with non-smooth hinge function form. Different from the MSE-type SBL, a cross entropy (CE)-type loss function named HRN that consists of a negative log-likelihood (NLL) term and a gradient penalty term with respect to inputs is proposed in \cite{OCC_NeurIPS2020_HRN}. OCC is indeed a binary classification obeying the Bernoulli distribution. However, the biggest difference from the binary classification is that the non-target samples are absent in training. The training of cross entropy solely on target samples may result in malformed learnable parameters. The gradient regularization is employed to alleviated this issue, but the training becomes harder. In addition, HRN is only verified on the simple multilayer perceptron (MLP).

In this paper, we follow the OCC design that seeks a hypersphere with the minimum radius to enclose the target data. Specifically, a novel logarithmic barrier function based OCC loss function (LBL) is first proposed by approximating the OCC objective smoothly. LBL assigns larger gradients to samples close to boundary, and thus achieves more compact hypersphere. However, the direct approximation by the logarithmic barrier function may cause the instability when samples lie on the boundary leading to the infinity loss. To address this issue, then, a unilateral relaxation Sigmoid function is introduced into LBL and a novel OCC loss named LBLSig is further proposed. Owing to the unilateral relaxation Sigmoid function, the optimization becomes smoother. Particularly, LBLSig can be seen as the fusion of MSE and CE. Experiments on various networks are conducted to verify the effectiveness of the proposed LBL/LBLSig. The results demonstrate the superior performance of our proposed algorithms.

\section{Brief Review }\label{sec:format}

The most common MSE-based OCC loss (MSE-OCL) is a simple but effective OCC loss function \cite{One_class_ELM, Dai_Haozhen_ML_OCELM, Tianlei_WSI_AE}.
Given an OCC training set with $N$ samples ${{\bf{x}}_i}$, MSE-OCL seeks a hypersphere centered at $\bf{c}$, that is
\begin{equation}\label{MSE_OCL}
\min_{ \boldsymbol{\mathcal{W}} } \quad  \frac{1}{N} \sum_{i=1}^{N} \left\| { {\boldsymbol{\Phi}} \left( {\bf{x}}_{i} ; {\boldsymbol{\mathcal{W}}}  \right) - {\bf{c}} } \right\|^{2} + \frac{{\lambda}}{2} \sum_{\ell=1}^{L}\left\|{\bf{W}}^{(\ell)}\right\|_{F}^{2},
\end{equation}
where ${\bf{c}}$ is the center of hypersphere, $\lambda$ is the trade-off parameter, and ${\boldsymbol{\Phi}} \left( {\bf{x}}_{i} ; {\boldsymbol{\mathcal{W}}}  \right)$ is the model output with the learnable parameter ${\boldsymbol{\mathcal{W}}}=\{ { {\bf{W}}^{(1)}, \cdots, {\bf{W}}^{(L)} } \}$. It can be readily seen that (\ref{MSE_OCL}) is a direct extension of MSE-based regression. In addition, the minimum-radius is obtained implicitly in MSE-OCL by minimizing the distance between the outputs of model ${\boldsymbol{\Phi}} \left( {\bf{x}}_{i} ; {\boldsymbol{\mathcal{W}}}  \right)$ and the center $\bf{c}$. Different from that, SBL \cite{Deep_SVDD} explicitly minimizes the radius of the hypersphere
\begin{equation}\label{Deep_SVDD_SBL}
\begin{aligned}
\min_{R, {\boldsymbol{\mathcal{W}}}} \quad & R^{2} + { \frac{{\lambda}_{1}}{N} } \sum_{i=1}^{N} \max \left\{ { 0, \left\| { {\boldsymbol{\Phi}} \left( {\bf{x}}_{i} ; {\boldsymbol{\mathcal{W}}}  \right) - {\bf{c}} } \right\|^{2} - R^{2} } \right\} \\
& +\frac{{\lambda}_2}{2} \sum_{\ell=1}^{L}\left\|{\bf{W}}^{(\ell)}\right\|_{F}^{2},
\end{aligned}
\end{equation}
where $\lambda_1$ and $\lambda_2$ are the trade-off parameters, $R$ is the radius of hypersphere. It is worth pointing out that the radius is updated in each mini-batch \cite{OCC_TNNLS_MOCCA_shit} by choosing a certain quantile from the sequence $\left\| { {\boldsymbol{\Phi}} \left( {\bf{x}}_{i} ; {\boldsymbol{\mathcal{W}}}  \right) - {\bf{c}} } \right\|^{2}$, as shown in the implementation\footnote{Code of Deep SVDD: \url{https://github.com/lukasruff/Deep-SVDD-PyTorch}.}. Therefore, SBL is the iterative hard sample mining in essence.

In contrast to the aforementioned MSE-type OCC losses, HRN \cite{OCC_NeurIPS2020_HRN} can be expressed as
\begin{equation}\label{HRN}
  \min_{\boldsymbol{\mathcal{W}}} \quad \underbrace{ -\sum_{i=1}^{N} { {\log\left( { {\text{Sig}}\left( {{\boldsymbol{\Phi}} \left( {\bf{x}}_{i} ; {\boldsymbol{\mathcal{W}}}  \right)} \right) } \right)} }  }_{\text{NLL}}
  +
  {\lambda}\underbrace{ \sum_{i=1}^{N}  { \left\| { {\nabla}_{\bf{x}}{{\boldsymbol{\Phi}} \left( {\bf{x}}_{i} ; {\boldsymbol{\mathcal{W}}}  \right)} } \right\|_F^{q} }  }_{\text{H-regularization}},
\end{equation}
where $\lambda$ and $q$ are two manual hyperparameters, and $\text{Sig}(\cdot)$ is the Sigmoid function that can be seen as the probability belonging to targets. It can be readily seen that minimizing NLL means to obtain high output value for target data, and the NLL term is essentially the direct extension from CE. H-regularization term is introduced as a constraint over the growth on the value of ${{\boldsymbol{\Phi}} \left( {\bf{x}}  \right)}$ caused by the single target class.

\section{Proposed methods}
\subsection{LBL}

OCC aims to seek a hypersphere that encloses targets. Therefore, the objective of OCC can be expressed as
\begin{equation}\label{objective_OCC}
    D_i = Dis\left( { {\boldsymbol{\Phi}} ({\bf{x}}_i;{\boldsymbol{\mathcal{W}}} ) , {\bf{c}} } \right) \le R , {\ }  i = 1, \cdots , N,
\end{equation}
where $Dis(\cdot)$ represents a distance function such as the L2-norm distance, and $R$ is the radius of the hypersphere. Then, a loss function can be constructed by directly approximating the inequality as
\begin{equation}\label{LBL_with_indicator_function}
\begin{aligned}
    \min\limits_{\boldsymbol{\mathcal{W}}} {\ } & \frac{1}{N}\sum\limits_{i=1}^{N}{ I \left( {u_i} \right) + \frac{{\lambda}}{2} \sum_{\ell=1}^{L}\left\|{\bf{W}}^{(\ell)}\right\|_{F}^{2}},\\
    {\rm{s.}}{\ }{\rm{t.}}{\ } & {u_i} =  {D_i^2} - R^2,
\end{aligned}
\end{equation}
where $I(u)$ is the indicator function that is defined as
\begin{equation}\label{indicator_function}
  I(u) = \left\{
   \begin{aligned}
      0, {\ }& u \leq 0, \\
      \infty, {\ }& u>0.
   \end{aligned}
  \right.
\end{equation}
Obviously,  minimizing $I(u)$ can enforce ${\boldsymbol{\Phi}} ({\bf{x}}_i;{\boldsymbol{\mathcal{W}}} )$ to fall inside the hypersphere $({\bf{c}},{R})$.
However, (\ref{LBL_with_indicator_function}) is difficult to be optimized due to that $I(u)$ is not differentiable. To address this issue, LBL is proposed to approximate (\ref{indicator_function}) smoothly by the logarithmic barrier function \cite{Book_convex_optimization, 9463718} as
\begin{equation}\label{LBL}
\begin{aligned}
    \min\limits_{\boldsymbol{\mathcal{W}}} {\ } &  - \frac{1}{N} \sum\limits_{i=1}^{N}{ \log \left( -{u_i} \right) + \frac{{\lambda}}{2} \sum_{\ell=1}^{L}\left\|{\bf{W}}^{(\ell)}\right\|_{F}^{2}},\\
    {\rm{s.}}{\ }{\rm{t.}}{\ } & {u_i} =  {D^2_i} - R^2.
\end{aligned}
\end{equation}
It can be seen that (\ref{LBL}) is convex and nondecreasing with respect to $u$. Specially, it increases to $\infty$ smoothly as $u$ increases to $0$.
In addition, the value of $I(u)$ and its gradient with respect to $u < 0$ are both equal to $0$ leading to the sample distributions inside the hypersphere unchange. 
On the contrary, Denote by $J_1({\boldsymbol{\mathcal{W}}})=- \frac{1}{N} \sum\limits_{i=1}^{N}{ \log \left( -{u_i} \right)}$. The gradient with respect to ${\boldsymbol{\mathcal{W}}}$ can be computed as
\begin{equation}\label{LBL_gradient}
{\frac{\partial{J_1}}{\partial{\boldsymbol{\mathcal{W}}}}} = {\frac{1}{N}} \sum_{i=1}^{N}{ {\frac{1}{-u_i}} \cdot {\frac{\partial{D^2_i}}{\partial{\boldsymbol{\mathcal{W}}}}} }.
\end{equation}
It can be seen that different penalties are assigned to target samples in LBL. The samples close to the boundary of the hypersphere (${Dis\left( { {\boldsymbol{\Phi}} ({\bf{x}}_i;{\boldsymbol{\mathcal{W}}} ) , {\bf{c}} } \right)} \to R$) will obtain higher penalty as well as the larger gradient value. In this way, LBL may pay more attention to contract these samples to center ${\bf{c}}$. Those samples away from the boundary that have small loss and gradient value are fine-tuned in the optimization of LBL for better distributions. 
However, this advantage may lead to vanishing gradients when fixing hypersphere radius $R$ during training.
To address this issue, $R$ is reset to twice the maximum distance when every several epochs. In this way, LBL is optimized by contracting the hypersphere with radius $R$ and thus keeps a fast learning speed. For the hypersphere center $\bf{c}$, it depends on models and will be given in  experiments.

\subsection{LBLSig}

The samples away from the boundary can obtain larger gradients in LBL than those close to center $\bf{c}$, which is similar with SBL but LBL is smoother. However, both SBL and LBL may suffer from the noises that always obtain larger gradients leading to wrong learning. To address this issue, a simple method is to slack the boundary by rejecting a percentage of targets samples, i.e., the radius $R$ is computed by
\begin{equation}\label{LBL_slack_radius}
R = {{Quantile}}\left( {Dis\left( { {\boldsymbol{\Phi}} ({\bf{x}}_i;{\boldsymbol{\mathcal{W}}} ) , {\bf{c}} } \right)},q \right),
\end{equation}
where ${Quantile}(\cdot,q)$ computes the $q$-th quantile of the sequence $\{ {Dis\left( { {\boldsymbol{\Phi}} ({\bf{x}}_i;{\boldsymbol{\mathcal{W}}} ) , {\bf{c}} } \right)},i=1,\cdots,N \}$. However, LBL becomes instability especially when samples lie on the boundary and thus fails to solve this situation.

Therefore, a smooth and robust OCC loss function named LBLSig is further developed, which can be expressed as
\begin{equation}\label{LBLSig}
\begin{aligned}
    \min\limits_{\boldsymbol{\mathcal{W}}} {\ } &  - \frac{1}{N} \sum\limits_{i=1}^{N}{ \log \left( {g(-{u_i})} \right) + \frac{{\lambda}}{2} \sum_{\ell=1}^{L}\left\|{\bf{W}}^{(\ell)}\right\|_{F}^{2}},\\
    {\rm{s.}}{\ }{\rm{t.}}{\ } & {u_i} =  {D^2_i} - R^2.
\end{aligned}
\end{equation}
Here,
\begin{equation}
  g(-u_i) = \left\{
   \begin{aligned}
      {\text{Sig}(-{u_i})}, {\ }& u_i \leq Q, \\
      {\text{Sig}(-{Q})}, {\ }& u_i > Q,
   \end{aligned}
  \right.
\end{equation}
where $Q>0$ is a relaxation hyperparameter that controls the tolerance of noisy samples. For $u_i > Q$, the Sigmoid function is truncated and the corresponding gradients are equal to $0$. Therefore, noisy samples can be filtered. For $u_i \leq Q$, denote by $v_i = {\text{Sig}}(-u_i)$ and $J_2({\boldsymbol{\mathcal{W}}})=- \frac{1}{N} \sum\limits_{i=1}^{N}{ \log \left( {\text{Sig}(-{u_i})} \right)}$. The gradient of $J_2({\boldsymbol{\mathcal{W}}})$ with respect to ${\boldsymbol{\mathcal{W}}}$ can be derived as
\begin{equation}\label{LBLSig_gradient}
\begin{aligned}
{\frac{\partial{J_2}}{\partial{\boldsymbol{\mathcal{W}}}}}
&= \sum_{i=1}^{N}{ {\frac{\partial{J_2}}{\partial{v_i}}}
{\frac{\partial{v_i}}{\partial{u_i}}}
{\frac{\partial{u_i}}{\partial{\boldsymbol{\mathcal{W}}}}} }\\
&= {\frac{1}{N}} \sum_{i=1}^{N}{
 {\frac{e^{u_i}}{1 + e^{u_i}}} {\frac{\partial{D^2_i}}{\partial{\boldsymbol{\mathcal{W}}}}} }\\
&= {\frac{1}{N}} \sum_{i=1}^{N}{ \left( {1-v_i} \right){\frac{\partial{D^2_i}}{\partial{\boldsymbol{\mathcal{W}}}}}  }.
\end{aligned}
\end{equation}
$v_i$ can be seen as the probability belonging to targets. Thus, samples close to center have large probability and thus small $\left( {1-v_i} \right)$, but the samples falling on the boundary and outside have small $v_i$ and thus the large gradient value. It is obvious that margin samples need to be assigned larger gradient for optimization than those close to the center $\bf{c}$.

Note that the proposed LBLSig is similar with the NLL term of HRN in (\ref{HRN}) but they are different in essence. The probability belonging to targets is obtained in HRN by mapping the model output to $[0,1]$ directly. H-regularization hence has to be used to limit the growth of the model output. Different from that, LBLSig minimizes ${Dis\left( { {\boldsymbol{\Phi}} ({\bf{x}}_i;{\boldsymbol{\mathcal{W}}} ) , {\bf{c}} } \right)}$ implicitly, which can be observed from (\ref{LBLSig_gradient}) that the gradient value relies on ${\frac{\partial{D^2_i}}{\partial{\boldsymbol{\mathcal{W}}}}}$ but the probability belonging to target $v_i$ is introduced to control the gradient. Particularly, LBLSig minimizes MSE-OCL implicitly when using L2-norm. Meanwhile, $J_2({\boldsymbol{\mathcal{W}}})$ is a obvious CE-type loss. Thus, LBLSig can be seen as the fusion of MSE and CE.

\section{Experiments}
\renewcommand{\arraystretch}{1.5}
\begin{table*}[ht]
\centering
\caption{Comparisons of AUCs with SBL and HRN on CIFAR10.}
\label{Comparisons_AUC_CIFAR10}
\resizebox{6.0in}{!}{
\begin{tabular}{cc|cccccccccc|c}
\hline
\multirow{2}{*}{Backbone} & \multirow{2}{*}{Methods} & 						\multicolumn{10}{c|}{Target Class} 				  			    			                                & \multirow{2}{*}{Avg} \\ 
																   \cline{3-12}
                          &                          & AIRPLANE   & AUTOMOBILE & BIRD        & CAT   	    & DEER       & DOG        & FROG  	   & HORSE      & SHIP        & TRUCK       & \\ 
\hline
\multirow{4}{*}{DSLeNet}  & SBL \cite{Deep_SVDD}     & 61.7       & 64.8       & 49.5        & 56.0  	  & 59.1       & 62.1       & 67.8       & 65.2       & 75.6        & 71.0       & 63.28 \\
                          & HRN                      & 64.73      & 64.05      & \bf{55.59}  & 51.81 	    & 62.99      & 50.06      & 64.13 	   & 53.53      & 71.51       & 66.8       & 60.52 \\
                          & LBL                      & 72.51      & \bf{71.00} & 54.82       & 56.79      & 63.04      & 60.24 	  & 58.86      & 71.92      & \bf{78.91}  & \bf{74.27} & 66.24 \\ 
                          & LBLSig                   & \bf{72.68} & 70.26      & 54.95       & \bf{58.81} & \bf{63.52} & \bf{64.47} & \bf{70.76} & \bf{72.25} & 78.89       & 74.25 	   & \bf{68.08} \\ 
\hline
\multirow{4}{*}{MLP}      & SBL                      & 76.43      & \bf{71.60} & 63.14 	     & 63.21 	  & 74.55      & 64.36      & 76.89 	 & 66.93      & \bf{83.45}  & 75.53 	 & 71.61 \\
                          & HRN                      & 58.80      & 62.27      & 51.80 	   & 54.94      & 51.01      & 53.13      & 56.44 	   & 53.02 	    & 60.00 	  & 63.84 	   & 56.53 \\
                          & LBL                      & 76.80      & 70.72      & \bf{66.52}  & \bf{63.66} & \bf{76.63} & \bf{65.16} & \bf{80.47} & 66.57 	    & 83.29 	  & \bf{77.83} & \bf{72.77}                \\
                          & LBLSig                   & \bf{76.81} & \bf{71.74} & 64.06 	   & 63.30 	    & 75.39      & 64.98 	  & 79.87 	   & \bf{66.97} & 83.41 	  & 77.28 	   & 72.38 \\ 
\hline
\multirow{4}{*}{OCITN}    & SBL                      & 66.17      & 64.32      & 60.40 	     & 54.21 	  & 61.00      & 58.74 	    & 66.43 	 & 65.06 	  & 75.22 	    & 68.88 	 & 64.04 \\
                          & HRN                      & N/A        & N/A        & N/A 	       & N/A 	    & N/A        & N/A   	  & N/A 	   & N/A 	    & N/A 	      & N/A 	   & N/A \\
                          & LBL                      & 76.38      & \bf{87.93} & 68.82       & 60.50 	    & 71.01      & \bf{72.17} & \bf{79.83} & 80.82 	    & 84.33       & 84.36 	   & 76.62 \\
                          & LBLSig                   & \bf{78.03} & 87.29      & \bf{69.42}  & \bf{60.91} & \bf{71.18}  & 71.99 	  & 78.54 	   & \bf{81.59} & \bf{84.80}  & \bf{84.96} & \bf{76.87}   \\
\hline
\end{tabular}
}
\end{table*}
\renewcommand{\arraystretch}{1}

\renewcommand{\arraystretch}{1.5}
\begin{table}[ht]
\caption{Comparisons of AUCs on non-image datasets.}
\label{Comparisons_AUC_nonimage}
\centering
\resizebox{2.8in}{!}{
\begin{tabular}{lcccc}
    \hline
    Dataset      & SBL               & HRN            & LBL          & LBLSig     \\
    \hline
    abalone      & 87.95             & 87.69          & 87.97        & \bf{88.03}        \\
    Arrhythmia   & 76.49             & 73.50          & 76.68        & \bf{77.17}        \\
    BASEHOCK     & 57.25             & 56.47          & 57.47        & \bf{59.82}        \\
    diabetes     & 72.47             & 69.77          & 71.91        & \bf{73.04}        \\
    Diabetic     & 63.36             & 62.74          & \bf{65.45}   & 63.68        \\
    ecoli        & 89.09             & 91.59          & \bf{96.24}   & 94.58        \\
    heart        & 83.51             & 82.39          & 82.72        & \bf{85.61}        \\
    leukemia     & 81.43             & 69.29          & \bf{83.69}   & 83.45        \\
    liver        & 58.79             & 61.25          & 58.65        & \bf{61.91}        \\
    magic        & 83.51             & 85.87          & \bf{86.93}   & 84.36        \\
    Online news  & 61.30             & 60.35          & 62.22        & \bf{62.37}        \\
    RCV1-4Class  & 78.27             & 54.84          & \bf{79.67}   & 79.05        \\
    Sonar        & 69.79             & 62.18          & 70.36        & \bf{72.73}   \\
    \hline
    \hline
    AVG          & 74.09             & 70.61          & 75.38        & \bf{75.83}        \\
    \hline
\end{tabular}
}

\end{table}
\renewcommand{\arraystretch}{1}

\subsection{Implementation details}

The proposed LBL and LBLSig are conducted on $4$ different networks for comparisons with SBL \cite{Deep_SVDD} and HRN \cite{OCC_NeurIPS2020_HRN}, including DSLeNet \cite{Deep_SVDD}, MLP with $2$ hidden layers, and OCITN \cite{OCC_IS_OCITN_shit}. For DSLeNet, the center $\bf{c}$ is set to be the mean of the outputs of the initialized DSLeNet on training data. For MLP, the mean of the training samples is computed as the hypersphere center $\bf{c}$. For OCITN, the center $\bf{c}$ is determined the same as \cite{OCC_IS_OCITN_shit}. 
The grid search method is employed to select the optimal hyperparameters for all algorithms. More detailed settings can be found in source code due to the page limitation.
In addition, SBL argues in \cite{Deep_SVDD} that networks with bias will lead to hypersphere collapse. Actually, SBL is non-convex and the network with bias leading to hypersphere collapse is a small probability. We hence do not add any constrains to the bias term in this paper.

Comparisons on $13$ non-image datasets and $1$ image dataset CIFAR10\footnote{\url{https://www.cs.toronto.edu/~kriz/cifar.html}} are conducted. The specification of non-image datasets can be found in \cite{Dai_Haozhen_OCC_MMCC, Tianlei_WSI_AE}, and the first class is set as the target and the remaining classes are regarded as outliers. For CIFAR10, each class is respectively chosen as the target class. The complete testing dataset is used to verify the model.
The area under the curve (AUC) of the receiver operating characteristic is used as evaluation metric.

\begin{table*}[!ht]
\caption{Comparisons of AUCs with the SOTA OCC algorithms.}\label{Comparisons_AUC_SOTA_CIFAR10}
\centering
\resizebox{4.5in}{!}{
\begin{threeparttable}
\begin{tabular}{lccccccc}
    \hline
    Target       & LSAR                    & OCGAN                    & HLS-OC                & MOCCA                         & HRN                         & LBL$^{*}$      & LBLSig$^{*}$         \\
    Class        & \cite{Deep_AD_AE_LSAR}  & \cite{Deep_AD_GAN_OCGAN} & \cite{Tianlei_WSI_AE} & \cite{OCC_TNNLS_MOCCA_shit}   & \cite{OCC_NeurIPS2020_HRN}                   &                &                \\
    \hline
    AIRPLANE     & 73.5                    & 75.7                     & 73.7                  & 66.0                          & 77.3                       & 76.38          & \textbf{78.03}         \\
    AUTOMOBILE   & 58.0                    & 53.1                     & 74.4                  & 74.6                          & 69.9                       & \textbf{87.93} & 87.29          \\
    BIRD         & 69.0                    & 64.0                     & 60.9                  & 57.5                          & 60.6                       & 68.82          & \textbf{69.42}          \\
    CAT          & 54.2                    & 62.0                     & 63.9                  & 60.1                          & \textbf{64.4}              & 60.50          & 60.91          \\
    DEER         & \textbf{76.1}           & 72.3                     & 71.0                  & 61.5                          & 71.5                       & 71.01          & 71.18          \\
    DOG          & 54.6                    & 62.0                     & 64.1                  & 68.4                          & 67.4                       & \textbf{72.17} & 71.99          \\
    FROG         & 75.1                    & 72.3                     & 78.8                  & 67.4                          & 77.4                       & \textbf{79.83} & 78.54          \\
    HORSE        & 53.5                    & 57.5                     & 70.6                  & 72.1                          & 64.9                       & 80.82          & \textbf{81.59}          \\
    SHIP         & 71.7                    & 82.0                     & 81.8                  & 79.2                          & 82.5                       & 84.33          & \textbf{84.80}         \\
    TRUCK        & 54.8                    & 55.4                     & 79.0                  & 77.3                          & 77.3                       & 84.36          & \textbf{84.96}          \\
    \hline
    AVG          & 64.05                   & 65.63                    & 71.82                 & 68.61                         & 71.32                      & 76.62          & \textbf{76.87}         \\
    \hline
\end{tabular}
\begin{tablenotes}
\item *Conducted on OCITN
\end{tablenotes}
\end{threeparttable}
}

\end{table*}

\subsection{Comparisons with SBL and HRN}

Table \ref{Comparisons_AUC_CIFAR10} gives the comparisons of AUCs on CIFAR10 under 3 different networks where the best results are highlighted in bold font.

\textbf{DSLeNet as backbone.} The results of SBL are taken from \cite{Deep_SVDD} for comparisons. The proposed LBLSig obtains the highest AUC value on average. Specially, LBLSig outperforms the SBL \cite{Deep_SVDD} on all classes and the increments are $10.98\%$ on AIRPLANE, $7.05\%$ on HORSE and $4.8\%$ on average. Similarly, the LBLSig generally has superior performance than HRN except the AUTOMOBILE class, and achieves more than $14.41\%$ increments on DOG, $18.72\%$ increments on HORSE and $7.56\%$ increments on average.

\textbf{MLP as backbone.} It can be readily seen that LBL and LBLSig obtain the first and second best performance on average, respectively. Particularly, LBL and LBLSig obtain much higher performance than HRN in our MLP structure. But it is worthy pointing out that the results reported in \cite{OCC_NeurIPS2020_HRN} are obtained using the structure of $\{3\times[1024$-$300]\}$-$[900$-$300]$-$[300$-$1]$ which is more complicated than our 2-hidden-layer MLP. The results in \cite{OCC_NeurIPS2020_HRN} are listed in Table \ref{Comparisons_AUC_SOTA_CIFAR10}, and it can be readily seen by comparisons that LBL and LBLSig generally outperform HRN of \cite{OCC_NeurIPS2020_HRN} and obtain better average AUC while only 2-hidden-layer MLP is used by us.

\textbf{OCITN as backbone.} As shown in Table \ref{Comparisons_AUC_CIFAR10}, LBLSig performs the best on 7 out of 10 classes and achieves the highest average AUC. On the contrary, the SBL on OCITN fails to obtain the promising performance and LBLSig has average $12.83\%$ improvements over SBL. In addition, HRN is originally required to fit a scalar, but OCITN aims to a color image, which leads to the hard optimization of HRN on OCITN. The results are hence fails to given in the table. However, it can be also said that LBL and LBLSig have wider applicability than HRN. In addition, the comparisons among networks show that LBL/LBLSig obtains the best performance when using OCITN as backbone. Particularly, LBLSig on OCITN achieves $4.49\%$ and $8.79\%$ increments over DSLeNet and MLP, respectively.

It should be pointed out that LBL may suffer from instability optimization and low robustness and thus LBLSig is developed to improve stability and robustness. LBLSig performs a litter better than LBL in experimental comparisons because the experiments are conducted on clean data such as CIFAR10. Due to the page limitation, the comparisons of robustness will demonstrated in future work. 

At last, the comparisons on non-image datasets are conducted and shown in Table \ref{Comparisons_AUC_nonimage}, where the single hidden layer MLP is used as the backbone. It can be readily seen that LBL/LBLSig outperform SBL and HRN on all non-image datasets and LBLSig wins the best average AUC value.

\subsection{Comparisons with SOTA algorithms}
Comparisons on CIFAR10 with common state-of-the-art (SOTA) OCC algorithms, including LSAR \cite{Deep_AD_AE_LSAR}, OCGAN \cite{Deep_AD_GAN_OCGAN}, HLS-OC \cite{Tianlei_WSI_AE}, MOCCA \cite{OCC_TNNLS_MOCCA_shit} and HRN \cite{OCC_NeurIPS2020_HRN}, are reported. OCITN is chosen as the backbone to conduct LBL/LBLSig. Results are listed in Tabel \ref{Comparisons_AUC_SOTA_CIFAR10} and it can be seen that the best performance is obtained by the proposed LBL/LBLSig. Specially, LBLSig achieves more than $12\%$, $10\%$, $4\%$, $8\%$ and $5\%$ increments of the average AUC over the LSAR, OCGAN, HLS-OC, MOCCA and HRN, respectively.

\section{Conclusions}

In this paper, two novel OCC loss functions named LBL and LBLSig were proposed. LBL employed the logarithmic barrier function to approximate the OCC objective smoothly. It assigned large gradients to margin samples and thus derived more compact hypersphere. LBLSig was further proposed to improve the optimization smoothness and stability by utilizing a unilateral relaxation Sigmoid function. LBLSig can be seen as the fusion of MSE and CE. The superior performance of the proposed LBL and LBLSig was experimentally verified in comparisons with several state-of-the-art OCC algorithms. 


 
%

\clearpage


\bibliographystyle{IEEEtran}
\bibliography{refs}

@ARTICLE{9774889,
  author={Zhang, Dasheng and Huang, Chao and Liu, Chengliang and Xu, Yong},
  journal={IEEE Signal Processing Letters}, 
  title={Weakly Supervised Video Anomaly Detection via Transformer-Enabled Temporal Relation Learning}, 
  year={2022},
  volume={29},
  number={},
  pages={1197-1201},
  doi={10.1109/LSP.2022.3175092}}

@ARTICLE{11208785,
  author={Wang, Sihan and Zhang, Shuhao and Cheng, Bo and Sheng, Shuwei},
  journal={IEEE Signal Processing Letters}, 
  title={An Anomalous Sound Detection Network Based on Time-Frequency Attention and Improved One-Class Softmax}, 
  year={2025},
  volume={32},
  number={},
  pages={4044-4048},
  doi={10.1109/LSP.2025.3624129}}

@ARTICLE{9585408,
  author={Ji, Haoran and Hou, Chunping and Yang, Yang and Fioranelli, Francesco and Lang, Yue},
  journal={IEEE Signal Processing Letters}, 
  title={A One-Class Classification Method for Human Gait Authentication Using Micro-Doppler Signatures}, 
  year={2021},
  volume={28},
  number={},
  pages={2182-2186},
  doi={10.1109/LSP.2021.3122344}}

@ARTICLE{9463718,
  author={Zhou, Jinyang and Li, Tiancheng and Wang, Xiaoxu and Zheng, Litao},
  journal={IEEE Signal Processing Letters}, 
  title={Target Tracking With Equality/Inequality Constraints Based on Trajectory Function of Time}, 
  year={2021},
  volume={28},
  number={},
  pages={1330-1334},
  doi={10.1109/LSP.2021.3090271}}

@inproceedings{transformaly_cvpr2022,
  title={Transformaly-Two (Feature Spaces) Are Better Than One},
  author={Cohen, Matan Jacob and Avidan, Shai},
  booktitle={Proceedings of the IEEE/CVF Conference on Computer Vision and Pattern Recognition},
  pages={4060--4069},
  year={2022}
}

@article{Deep_AD_SAE_OCSVM,
  author = {Andrews, Jerone TA and Morton, Edward J and Griffin, Lewis D},
  title = {{Detecting Anomalous Data Using Auto-Encoders}},
  journal = {International Journal of Machine Learning and Computing},
  volume = {6},
  number = {1},
  pages = {21},
  year = {2016}
}

@article{Deep_AD_DAE_OCSVM,
author = {Xu, Dan and Ricci, Elisa and Yan, Yan and Song, Jingkuan and Sebe, Nicu},
journal = {arXiv preprint arXiv:1510.01553},
title = {{Learning Deep Representations of Appearance and Motion for Anomalous Event Detection}},
year = {2015}
}

@article{Deep_AD_DBN_OCSVM,
author = {Erfani, Sarah M and Rajasegarar, Sutharshan and Karunasekera, Shanika and Leckie, Christopher},
journal = {Pattern Recognition},
pages = {121--134},
publisher = {Elsevier},
title = {{High-dimensional and large-scale anomaly detection using a linear one-class SVM with deep learning}},
volume = {58},
year = {2016}
}

@inproceedings{Deep_AD_ensemble_AEs_reconstruction,
author = {Chen, Jinghui and Sathe, Saket and Aggarwal, Charu and Turaga, Deepak},
booktitle = {Proceedings of 2017 SIAM International Conference on Data Mining},
pages = {90--98},
title = {{Outlier Detection with Autoencoder Ensembles}},
year = {2017}
}

@article{Deep_AD_VAE_reconstruction,
author = {An, Jinwon and Cho, Sungzoon},
journal = {Special Lecture on IE},
number = {1},
title = {{Variational Autoencoder based Anomaly Detection using Reconstruction Probability}},
volume = {2},
year = {2015}
}

@inproceedings{Deep_AD_AE_DAE_reconstruction,
author = {Sakurada, Mayu and Yairi, Takehisa},
booktitle = {Proceedings of MLSDA 2014 2nd Workshop on Machine Learning for Sensory Data Analysis},
pages = {4--11},
title = {{Anomaly Detection Using Autoencoders with Nonlinear Dimensionality Reduction}},
year = {2014}
}

@article{OCC_TNNLS_MOCCA_shit,
author = {Massoli, Fabio Valerio and Falchi, Fabrizio and Kantarci, Alperen and Akti, Seymanur and Ekenel, Hazim Kemal and Amato, Giuseppe},
journal = {IEEE Transactions on Neural Networks and Learning Systems},
number = {6},
pages = {2313--2323},
title = {{MOCCA: Multilayer One-Class Classification for Anomaly Detection}},
volume = {33},
year = {2022}
}

@article{OCC_IS_OCITN_shit,
author = {Hayashi, Toshitaka and Fujita, Hamido and Hernandez-Matamoros, Andres},
journal = {Information Sciences},
pages = {217--234},
title = {{Less complexity one-class classification approach using construction error of convolutional image transformation network}},
volume = {560},
year = {2021}
}

@inproceedings{OCC_CVPR_PANDA_supershit,
author = {Reiss, Tal and Cohen, Niv and Bergman, Liron and Hoshen, Yedid},
booktitle = {Proceedings of the IEEE/CVF Conference on Computer Vision and Pattern Recognition},
pages = {2806--2814},
title = {{PANDA: Adapting Pretrained Features for Anomaly Detection and Segmentation}},
year = {2021}
}

@inproceedings{OCC_NeurIPS2020_HRN,
author = {Hu, Wenpeng and Wang, Mengyu and Qin, Qi and Ma, Jinwen and Liu, Bing},
booktitle = {Advances in Neural Information Processing Systems},
pages = {19111--19124},
title = {{HRN: A Holistic Approach to One Class Learning}},
volume = {33},
year = {2020}
}

@inproceedings{OCC_GODS_OCSVM,
author = {Wang, Jue and Cherian, Anoop},
booktitle = {Proceedings of the IEEE International Conference on Computer Vision},
pages = {8201--8211},
title = {{GODS: Generalized One-class Discriminative Subspaces for Anomaly Detection}},
year = {2019}
}

@inproceedings{Deep_SVDD,
author = {Ruff, Lukas and Vandermeulen, Robert A and G{\"o}rnitz, Nico and Deecke, Lucas and Siddiqui, Shoaib Ahmed and Binder, Alexander and M{\"u}ller, Emmanuel and Kloft, Marius},
booktitle = {Proceedings of the International Conference on Machine Learning},
pages = {4390--4399},
title = {{Deep One-Class Classification}},
year = {2018}
}

@inproceedings{OCC_NIPS_Deep_Anomaly_Detection_GT,
author = {Golan, Izhak and El-Yaniv, Ran},
booktitle = {Proceedings of Advances in Neural Information Processing Systems},
title = {{Deep Anomaly Detection Using Geometric Transformations}},
volume = {31},
year = {2018}
}

@article{One_class_ELM,
author = {Leng, Qian and Qi, Honggang and Miao, Jun and Zhu, Wentao and Su, Guiping},
doi = {10.1155/2015/412957},
journal = {Mathematical Problems in Engineering},
title = {{One-Class Classification with Extreme Learning Machine}},
volume = {2015},
year = {2015}
}

@article{Dai_Haozhen_ML_OCELM,
author = {Dai, Haozhen and Cao, Jiuwen and Wang, Tianlei and Deng, Muqing and Yang, Zhixin},
doi = {10.1016/j.neunet.2019.03.004},
journal = {Neural Networks},
pages = {11--22},
title = {{Multilayer one-class extreme learning machine}},
volume = {115},
year = {2019}
}

@article{Tianlei_WSI_AE,
author = {Wang, Tianlei and Cao, Jiuwen and Lai, Xiaoping and Wu, Q. M.Jonathan},
doi = {10.1109/TNNLS.2020.3015860},
journal = {IEEE Transactions on Neural Networks and Learning Systems},
number = {8},
pages = {3770--3776},
title = {{Hierarchical One-Class Classifier with Within-Class Scatter-Based Autoencoders}},
volume = {32},
year = {2021}
}

@article{Dai_Haozhen_OCC_MMCC,
author = {Wang, Tianlei and Cao, Jiuwen and Dai, Haozhen and Lei, Baiying and Zeng, Huanqiang},
doi = {10.1109/MIS.2021.3122958},
journal = {IEEE Intelligent Systems},
pages = {1--7},
title = {{Robust Maximum Mixture Correntropy Criterion based One-Class Classification Algorithm}},
year = {2021}
}

@article{SVDD,
  author={Tax, David MJ and Duin, Robert PW},
  title={Support vector data description},
  journal={Machine Learning},
  volume={54},
  number={1},
  pages={45-66},
  year={2004},
  publisher={Springer}
}

@book{Book_convex_optimization,
  title={{Convex Optimization}},
  author={Boyd, Stephen and Vandenberghe, Lieven},
  year={2004},
  publisher={Cambridge university press}
}

@article{review_DL_AD,
  title={Deep learning for anomaly detection: A review},
  author={Pang, Guansong and Shen, Chunhua and Cao, Longbing and Hengel, Anton Van Den},
  journal={ACM Computing Surveys (CSUR)},
  volume={54},
  number={2},
  pages={1--38},
  year={2021},
  publisher={ACM New York, NY, USA}
}

@article{OCSVM,
  author={Sch{\"o}lkopf, Bernhard and Platt, John C and Shawe-Taylor, John and Smola, Alex J and Williamson, Robert C},
  title={Estimating the support of a high-dimensional distribution},
  journal={Neural Computation},
  volume={13},
  number={7},
  pages={1443-1471},
  year={2001},
  publisher={MIT Press}
}

@inproceedings{Deep_AD_AE_LSAR,
  title={Latent Space Autoregression for Novelty Detection},
  author={Abati, Davide and Porrello, Angelo and Calderara, Simone and Cucchiara, Rita},
  booktitle={Proceedings of IEEE Conference on Computer Vision and Pattern Recognition},
  pages={481-490},
  year={2019}
}

@article{Deep_AD_AE_pidhorskyi2018generative,
  title={Generative probabilistic novelty detection with adversarial autoencoders},
  author={Pidhorskyi, Stanislav and Almohsen, Ranya and Doretto, Gianfranco},
  journal={Advances in neural information processing systems},
  volume={31},
  year={2018}
}

@inproceedings{Deep_AD_AE_zhou2017anomaly,
  title={Anomaly detection with robust deep autoencoders},
  author={Zhou, Chong and Paffenroth, Randy C},
  booktitle={Proceedings of the 23rd ACM SIGKDD international conference on knowledge discovery and data mining},
  pages={665--674},
  year={2017}
}

@inproceedings{Deep_AD_AE_gong2019memorizing,
  title={Memorizing normality to detect anomaly: Memory-augmented deep autoencoder for unsupervised anomaly detection},
  author={Gong, Dong and Liu, Lingqiao and Le, Vuong and Saha, Budhaditya and Mansour, Moussa Reda and Venkatesh, Svetha and Hengel, Anton van den},
  booktitle={Proceedings of the IEEE/CVF International Conference on Computer Vision},
  pages={1705--1714},
  year={2019}
}

@article{Deep_AD_AE_wang2020advae,
  title={adVAE: A self-adversarial variational autoencoder with Gaussian anomaly prior knowledge for anomaly detection},
  author={Wang, Xuhong and Du, Ying and Lin, Shijie and Cui, Ping and Shen, Yuntian and Yang, Yupu},
  journal={Knowledge-Based Systems},
  volume={190},
  pages={105187},
  year={2020},
  publisher={Elsevier}
}

@inproceedings{Deep_AD_GAN_AnoGAN,
author = {Schlegl, Thomas and Seeb{\"o}ck, Philipp and Waldstein, Sebastian M and Schmidt-Erfurth, Ursula and Langs, Georg},
booktitle = {Proceedings of International Conference on Information Processing in Medical Imaging},
pages = {146--157},
title = {{Unsupervised Anomaly Detection with Generative Adversarial Networks to Guide Marker Discovery}},
year = {2017}
}

@inproceedings{Deep_AD_GAN_AnoGAN_ALOCC,
  title={Adversarially learned one-class classifier for novelty detection},
  author={Sabokrou, Mohammad and Khalooei, Mohammad and Fathy, Mahmood and Adeli, Ehsan},
  booktitle={Proceedings of the IEEE conference on computer vision and pattern recognition},
  pages={3379--3388},
  year={2018}
}

@inproceedings{Deep_AD_GAN_AnoGAN_OGN,
  title={Old is gold: Redefining the adversarially learned one-class classifier training paradigm},
  author={Zaheer, Muhammad Zaigham and Lee, Jin-ha and Astrid, Marcella and Lee, Seung-Ik},
  booktitle={Proceedings of the IEEE/CVF Conference on Computer Vision and Pattern Recognition},
  pages={14183--14193},
  year={2020}
}

@inproceedings{Deep_AD_GAN_abati2019latent,
  title={Latent space autoregression for novelty detection},
  author={Abati, Davide and Porrello, Angelo and Calderara, Simone and Cucchiara, Rita},
  booktitle={Proceedings of the IEEE/CVF conference on computer vision and pattern recognition},
  pages={481--490},
  year={2019}
}

@inproceedings{Deep_AD_GAN_sabokrou2018adversarially,
  title={Adversarially learned one-class classifier for novelty detection},
  author={Sabokrou, Mohammad and Khalooei, Mohammad and Fathy, Mahmood and Adeli, Ehsan},
  booktitle={Proceedings of the IEEE conference on computer vision and pattern recognition},
  pages={3379--3388},
  year={2018}
}

@inproceedings{Deep_AD_GAN_OCGAN,
  title={{OCGAN}: One-class novelty detection using gans with constrained latent representations},
  author={Perera, Pramuditha and Nallapati, Ramesh and Xiang, Bing},
  booktitle={Proceedings of IEEE Conference on Computer Vision and Pattern Recognition},
  pages={2898--2906},
  year={2019}
}

@article{Deep_AD_GAN_schlegl2019f,
  title={f-AnoGAN: Fast unsupervised anomaly detection with generative adversarial networks},
  author={Schlegl, Thomas and Seeb{\"o}ck, Philipp and Waldstein, Sebastian M and Langs, Georg and Schmidt-Erfurth, Ursula},
  journal={Medical image analysis},
  volume={54},
  pages={30--44},
  year={2019},
  publisher={Elsevier}
}



 





\end{document}